# Legendre Memory Unit with A Multi-Slice Compensation Model for Short-Term Wind Speed Forecasting Based on Wind Farm Cluster Data

Mumin Zhang, Haochen Zhang, Xin Zhi Khoo, Yilin Zhang, Nuo Chen, Ting Zhang, Junjie Tang

*Abstract* — With more wind farms clustered for integration, the short-term wind speed prediction of such wind farm clusters is critical for normal operation of power systems. This paper focuses on achieving accurate, fast, and robust wind speed prediction by full use of cluster data with spatial-temporal correlation. First, weighted mean filtering (WMF) is applied to denoise wind speed data at the single-farm level. The Legendre memory unit (LMU) is then innovatively applied for the wind speed prediction, in combination with the Compensating Parameter based on Kendall rank correlation coefficient (CPK) of wind farm cluster data, to construct the multi-slice LMU (MSLMU). Finally, an innovative ensemble model WMF-CPK-MSLMU is proposed herein, with three key blocks: data pre-processing, forecasting, and multi-slice compensation. Advantages include: 1) LMU jointly models linear and nonlinear dependencies among farms to capture spatial-temporal correlations through backpropagation; 2) MSLMU enhances forecasting by using CPK-derived weights instead of random initialization, allowing spatial correlations to fully activate hidden nodes across clustered wind farms.; 3) CPK adaptively weights the compensation model in MSLMU and complements missing data spatially, to facilitate the whole model highly accurate and robust. Test results on different wind farm clusters indicate the effectiveness and superiority of proposed ensemble model WMF-CPK-MSLMU in the short-term prediction of wind farm clusters compared to the existing models.

## I. INTRODUCTION

FULL utilization of the clean and renewable energy is an important trend in human sustainable development, in front of the deteriorating environmental pollution caused by massive usage of fossil fuels [1]. Amongst the renewable energy sources, the utilization of wind energy has increased significantly in the past decade. In 2021, the global wind energy capacity was 830 GW, and then the percentage increased by 9.15% in 2022 (906 GW) and 12.69% in 2023 (1021 GW) [2]. In addition, wind power generation is more common for the wind farm clusters composed of multiple wind farms. For example, Hornsea Wind Farm in the UK, one of the largest offshore wind farms, is divided into three parts, whose actual output reaches 800 MW [3]. For onshore wind farms, one of the largest wind power plants in the world, Jiuquan Wind Power Base in Gansu, China covers an area of thousands of square kilometers. It reached a total capacity at the 10 GW level and is continuing to increase the number of units, as well as expanding its scale [4]. Amongst these large wind farms, they are generally divided into different groups for operation according to the terrain, which means that they are distributed and managed in clusters [4].

Site selection for establishing a wind farm generally requires various factors to be considered, including wind speed, wind direction, and geographical location [5], [6]. To guarantee the security and availability of wind power generation, wind speed forecasting (WSF) is often regarded as an indispensable task in practical power system operation. Accurate wind speed forecasting has long been a critical challenge for the integration of large-scale wind energy [7].

Due to the random, fluctuating, and intermittent nature of wind resources, WSF is difficult to achieve accurate values [6]. To address these issues, much research has focused on single wind farm prediction. Statistical methods use historical data to build probabilistic links between predictions and measurements, avoiding the need to model complex physical mechanisms [8]. Time series methods, for example, auto-regressive and moving average (ARMA), new shape-position comprehensive evaluation, etc. have achieved good results in short-term WSF scenarios [9], [10]. Machine learning methods, including artificial neural networks (ANN), back propagation neural networks (BP) and extreme learning machines (ELM), use a large amount of historical data to fit the non-smoothness of the wind speed series by analyzing the intrinsic relationship throughout historical data to achieve a high prediction accuracy [11]-[13]. Deep learning methods, for instance, convolutional neural networks (CNN) or long short-term memory (LSTM) in recurrent neural network (RNN) research on the applicability to further explore the wind data characteristics. Additionally, some models for wind speed forecasting have been developed

This work was supported by the National Natural Science Foundation of China (Grant No. 52177071). (Corresponding author: Junjie Tang.)

Mumin Zhang is with the Daniel J. Epstein Department of Industrial and Systems Engineering, Viterbi School of Engineering, University of Southern California, Los Angeles, CA 90089, USA (2024-2026). (e-mail: muminzha@usc.edu).

Haochen Zhang is with the Department of Computer Science, Rice University, Houston, USA (email: hz112@rice.edu).

Xin Zhi Khoo is with the Department of Ecology & Evolutionary Biology, University of California, Irvine, CA 92697, USA (email: xkhoo@uci.edu).

Junjie Tang and Yilin Zhang are with the State Key Laboratory of Power Transmission Equipment Technology, School of Electrical Engineering, Chongqing University, Chongqing 400044, China (emails: zhangyilin@stu.cqu.edu.cn; chennuo@cqu.edu.cn; tangjunjie@cqu.edu.cn).

Nuo Chen is with the Jiangsu Electric Power Co. Ltd Suzhou Branch, State Grid Corporation of China, Suzhou 215004, China (emails: norachen_0w0@163.com).

Ting Zhang is with the School of Mathematical Sciences, University of Science and Technology of China, University of Science and Technology of China, Hefei 230022, China (email: zhangting@mail.ustc.edu.cn).

utilizing the deep belief network (DBN) [14]-[16].

However, based on the above studies, it can be found that most of them focused on the accuracy of prediction in a single wind farm while the importance of physical factors such as the wind direction, temperature, wake effect and also terrain were neglected. These factors can be reflected in the historical wind speed of adjacent wind farms, which indicates that the WSF for multiple wind farms based on cluster data is more reasonable than that for a single wind farm [17].

So, utilizing the data characteristics of wind farm groups is presented in wind power prediction (WPP) due to this reason. A power prediction of a wind farm cluster, which is based on LSTM with the spatiotemporal characteristics extraction on CNN, considers the influences of wind speed, wind direction, and temperature on the power output of wind farm clusters [18]. Moreover, for data mining based on a large number of samples, a new multi-feature similarity matching (MFSM) method is proposed in order to improve the efficiency and accuracy of power prediction of wind farm clusters [19]. Regarding WSF, one method is to increase the dimensionality of input data by constructing the input sets, thus increasing the complexity and accuracy of prediction models, while the construction of input sets is determined by correlation. For instance, the predictive spatio-temporal network (PSTN) and the predictive deep convolutional neural network (PDCNN) proposed by Qiaomu Zhu et al. promote the research on the WSF of multiple wind turbines, providing a general mathematical description of the problem, and establishing it as a space-time series prediction problem [17],[20]. The former one builds a unified framework integrating a CNN and LSTM [17], while the latter combines CNNs with a multi-layer perceptron (MLP) [20]. Starting from the nature of space-time sequence, the common point of the two methods is to propose a two-stage modeling strategy, which first extracts spatial features and then establishes the temporal relationships [17]. Although Zhu mentioned that the spatial-temporal correlation is not only applicable to several wind turbines within the same wind farm, but also to the multiple wind farms or wind farm clusters [20]; however CNN, which is mainly for image processing methods of turbines in a single wind farm, is not applicable for the situation in the long distance of different wind farms and their irregular arrangement. So, further studies need to be explored.

Therefore, it is worth trying to employ the data of the wind farm cluster to improve the WSF accuracy of each single wind farm. Legendre memory unit (LMU), a new RNN-based variant, which is advantageous for prediction efficiency and handling large-capacity data [21], is first applied to the field of WSF in this paper. Moreover, to improve its accuracy, the multi-slice Legendre memory unit (MSLMU) is initially proposed.

A suitable model is required to efficiently use wind speed data for prediction. However, random fluctuations often arise, especially when data is missing continuously due to issues like communication failure or device damage. To address this problem, interpolation methods are adopted, typically from these two perspectives, spatially and temporally [22]. For example, a new method called vertically correlated echelon model (VCEM), which uses the vertical correlation of wind speeds at different heights to fill in such missing data, has achieved a significant improvement in the prediction accuracy than traditional methods like artificial neural network with an adaptive neuro-fuzzy (ANFIS), the revised power law and 1/7 power law [23]. Temporally, a common method is the moving average approach (MAA), which makes up for the missing data by means of calculating the average value of the previous amount of data with the defined autoregressive order [22].

As both are single-farm forecasting methods, we refer to them as single-farm methods to contrast with our cluster-farm approach. VCEM is restricted by the necessity to obtain the data of different heights and MAA has poor accuracy due to the only utilization of historical data in a linear way. Therefore, to adequately consider the spatial-temporal correlation and hence conduct an accurate prediction, a novel cluster-farm approach based on the compensating parameter determined by the Kendall rank correlation coefficient (CCK) is first proposed in this paper.

When LMU and CCK are combined for application, both spatial and temporal properties of the wind speed data can be fully characterized and result in a significant improvement in accuracy for both complete data and missing data conditions.

In this paper, an innovative ensemble method known as Weighted Mean Filter (WMF) – compensating parameter that uses the KRCC (CPK) – MSLMU is proposed for short-term wind speed forecasting. While the forecasting target remains a single wind farm (or each farm within a cluster), our approach uniquely leverages cluster-wide data, addressing a common limitation in the literature where models predict individual farm speeds using only their own data. Specifically, the WMF cleans white noise from the series, Kendall-based compensating parameter (CPK) corrects inputs using spatial correlation, and MSLMU enhances the WSF model. The main contributions are:

(1) LMU, a novel RNN architecture combining linear and nonlinear units, is applied to wind speed forecasting (WSF) for the first time, which effectively models nonlinear wind patterns and outperforms many deep learning baselines.

(2) To enhance the forecast accuracy of each farm within a cluster, MSLMU augments LMU by replacing random weight initialization with CPK-determined weights, exploiting spatial correlation to activate all hidden nodes.

(3) The CPK method significantly improves the robustness: even with varying lengths of data loss, WMF-CPK-MSLMU maintains high accuracy, whereas single-farm methods degrade rapidly.

The structure of the remaining framework of this paper is as follows: Section II introduces the theory of LMU. Section III illustrates the WMF-CPK-MSLMU model in detail. Section IV presents five aspects of case studies, and the test results are discussed comprehensively. Section V draws the conclusions.

## II. THEORY OF LMU FOR PREDICTION

### A. Introduction of Legendre Memory Unit (LMU)

WSF belongs to a typical time series regression problem. Traditional forward fully connected neural networks, in general, cannot efficiently learn the influence of historical information about a time series on its future variation trend. The recurrent neural network is a basic model that adds its output-to-input recurrent structure to the typical implicit layer neurons to recall historical information about time series [24]. It can efficiently link contextual data, especially in time series modeling [25].

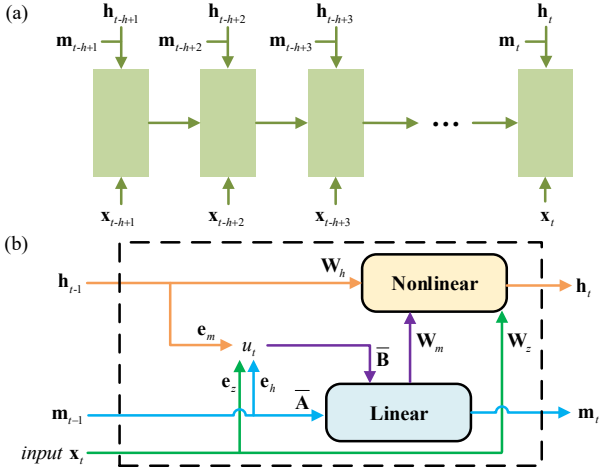

Fig. 1. The illustrations on the structures of LMU: (a) structure of the LMU layers; (b) memory cell of LMU

As the research progresses, some new RNN-based variants are promoted and developed. LMU is a promising type of neural regenerative memory cell that belongs to a class of RNN units. It can maintain information across time windows dynamically, which consists of coupled linear and nonlinear systems. Among them, the linear system is further referred to as a delayed network (DN) in the algorithm, which can project the input signal onto the Legendre base in a sliding window [26]. Taking DN as internal memory and using nonlinear recursive layers to calculate the functions in time, LMU outperforms LSTM and other RNN algorithms in terms of efficiency [21]. This good ability of LMU to deal with the nonlinear data characteristics is fundamental that enables it to be feasible in the field of WSF.

*B. Mathematical Modeling of LMU*

The layers in Fig. 1(a) give an overview of the high-level structure of the LMU operation. Specifically, for a particular layer, as shown in Fig. 1(b), the main component of the LMU is a memory cell. It can orthogonalize the history information of the input signal over a continuous time period through a sliding window with length $\theta$ [21]. The output of the unit will be derived from the time delay transfer function $G(s)$, which is approximated by the ordinary differential equation (ODE).

$$\frac{dm_i}{dt} - A_{ij} m_i(t) = B_i u_i(t) \tag{1}$$

$$G(s) = e^{-\theta s} \tag{2}$$

where $u_i$ and $m_i$ are the $i^{th}$ element of the input signal for the linear function (**u**) and the memory cell (**m**) with $d$ dimensions. The ideal state matrix **A** and vector **B** can be handled through Padé approximants to obtain the corresponding approximation in what follows [21]:

$$\mathbf{A} \in \mathbf{R}^{d \times d}, A_{ij} = \frac{2i+1}{\theta} \begin{cases} -1 & i < j \\ (-1)^{i-j+1} & i \geq j \end{cases} \tag{3}$$

$$\mathbf{B} \in \mathbf{R}^{d \times 1}, B_i = \frac{2i+1}{\theta}(-1)^i \tag{4}$$

For the obtained ordinary differential equation through an approximation, given **m** in the equation (1), it can be decoded using the shifted Legendre polynomial as below:

$$u(t-\theta') \approx \sum_{i=0}^{d-1} P_i\left(\frac{\theta'}{\theta}\right) m_i(t),\ 0 \leq \theta' \leq \theta \tag{5}$$

$$P_i(r) = (-1)^i \sum_{j=0}^{i} \binom{i}{j}\binom{i+j}{j}(-r)^j \tag{6}$$

where $P_i$ is exactly the Legendre polynomial representing the $i^{th}$ shift. For the RNNs, discretization is required for the ordinary differential equations at discrete time instants.

$$\mathbf{m}_t = \bar{\mathbf{A}} \mathbf{m}_{t-1} + \bar{\mathbf{B}} u_t \tag{7}$$

$$\bar{\mathbf{A}} = \left(\frac{\Delta t}{\theta}\right) \mathbf{A} + \mathbf{I} \tag{8}$$

$$\bar{\mathbf{B}} = \left(\frac{\Delta t}{\theta}\right) \mathbf{B} \tag{9}$$

where $\Delta t$ corresponds to the window length $\theta$, $(\bar{\mathbf{A}}, \bar{\mathbf{B}})$ is the discretized form of $(\mathbf{A}, \mathbf{B})$. Combining the above information, the state update equation of LMU can be finally obtained as:

$$u_t = \mathbf{e}_x^T \mathbf{x}_t + \mathbf{e}_h^T \mathbf{h}_{t-1} + \mathbf{e}_m^T \mathbf{m}_{t-1} \tag{10}$$

$$\mathbf{h}_t = f(\mathbf{W}_x \mathbf{x}_t + \mathbf{W}_h \mathbf{h}_{t-1} + \mathbf{W}_m \mathbf{m}_t) \tag{11}$$

where $\mathbf{x}_t$ is the input at moment $t$, $\mathbf{h}_t$ is the $n$-dimensional state vector, $\mathbf{m}_t$ is the $d$-dimensional memory vector, $\mathbf{e}_x$, $\mathbf{e}_h$, $\mathbf{e}_m$ are the learned encoding vectors, $\mathbf{W}_x$, $\mathbf{W}_h$, $\mathbf{W}_m$ are the learned kernels, and $f$ is a specific nonlinear form (e.g., tanh). The state depends on the input received, the prior condition (e.g., $\mathbf{h}_{t-1}$) and the present memory. A single LMU unit decouples the hidden states from the memory states, and thus only $(n+d)$ memory variables need to be saved during the time steps [21].

Therefore, LMU decouples the linear storage units and the nonlinear hidden state units, which takes the advantage of comprehending the interrelation between an optimal linear dynamical memory system and a nonlinear function. Moreover, by use of non-saturated storage units, LMU makes it possible to converge quickly and perform efficiently [27].

## III. An Ensemble Model WMF-CPK-MSLMU for Wind Speed Prediction

*A. Multi-slice Legendre Memory Unit*

When applied as the method for a multi-slice compensation model, MSLMU, which is firstly presented in this paper, relieves the disadvantage of LMU, low single layer accuracy; and enhances high efficiency and large data processing capacity.

As shown in Fig. 2, MSLMU contains $n$ slices and each LMU unit varies from labels. Correspondingly, slice 1 to slice $n$ are introduced as follows: take wind farm 1 as an example.

Slice 1: The historical wind speed of wind farm 1 is used as input of the first slice of LMU to obtain the predicted wind speed $pred_1$. $error_1$ denotes the difference between the real value and the predicted value in the first slice.

Slice 2: The historical wind speed of wind farm 2 is used as input for the second slice of LMU, where the training label is $error_1$ from the first slice. $error_2$ denotes the difference between $error_1$ and $pred_2$.

…

*Slice n:* The training label is $error_{n-1}$, propagated from the previous slice. The output prediction is $pred_n$, and the residual error is computed as $error_n$

Integrated with the above information, the prediction flow equation of the MSLMU model can be built as follows:

$$pred_{plus} = \sum_{i=1}^{n} pred_i \qquad (12)$$

where $pred_{plus}$ is the modified prediction value of the WSF in MSLMU for wind farm 1, whilst $pred_1$, $pred_2$, …, and $pred_n$ are the prediction results of each slice, respectively.

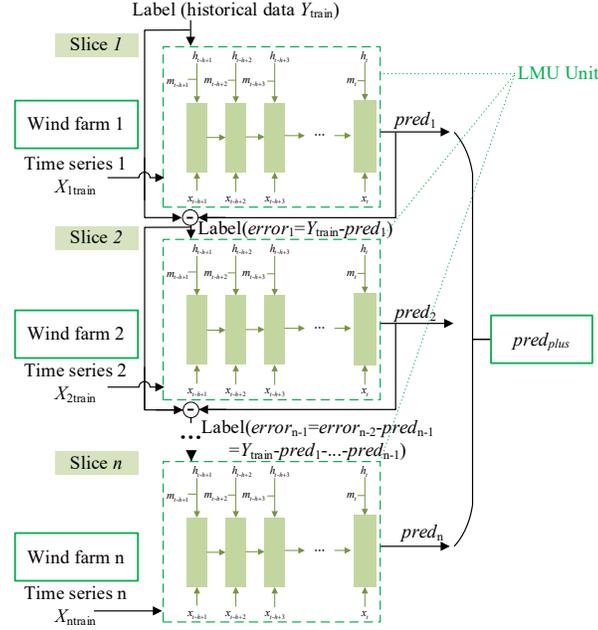

Fig. 2. Process of the training set for MSLMU

### B. Compensating Parameter Based on the KRCC (CPK)

The Kendall rank correlation coefficient is a statistical value adopted to measure the correlation of two random variables. A Kendall test is a parameter-free hypothesis test that uses the calculated rank correlation coefficient to quantify the statistical dependence of two random variables [28]. It has the advantages of monotonically increasing transformation invariance such as the monotone increasing transformation invariance, robustness, and a broader interpretation of the correlation, which cannot be found in other correlation coefficient methods [28], [29].

The correlation functions can be characterized and updated periodically, although the time series show no evident cyclic behavior [30]. According to this principle, the wind speed data at the same moment of each day are collected into series, namely periodic time series (PTS). For example, as depicted in Fig. 3, *x*, including, $x_1, x_2,..., x_i, ..., x_n$, is the PTS for wind farm 1 from the 1st day to the $m^{th}$ day at 00:00.

The calculation process of KRCC among PTSs is shown in Fig. 3 with red arrows, which is defined as follows [28]:

$$\xi(x_i, x_j, y_i, y_j) = \begin{cases} 1 & \text{if } (x_i - x_j)(y_i - y_j) > 0 \\ 0 & \text{if } (x_i - x_j)(y_i - y_j) = 0 \\ -1 & \text{if } (x_i - x_j)(y_i - y_j) < 0 \end{cases} \qquad (13)$$

$$r = \frac{2\sum_{i=1}^{n'-1} \sum_{j=i+1}^{n'} \xi(x_i, x_j, y_i, y_j)}{n'(n'-1)} \qquad (14)$$

where $r$ is the KRCC; $\xi$ stands for correlation characteristics of any two pairs of wind speed series; $n'$ is the length of the time sequence. The range of KRCC is from -1 to 1, where -1 stands for the negative correlation, 1 for the positive, and 0 for none.

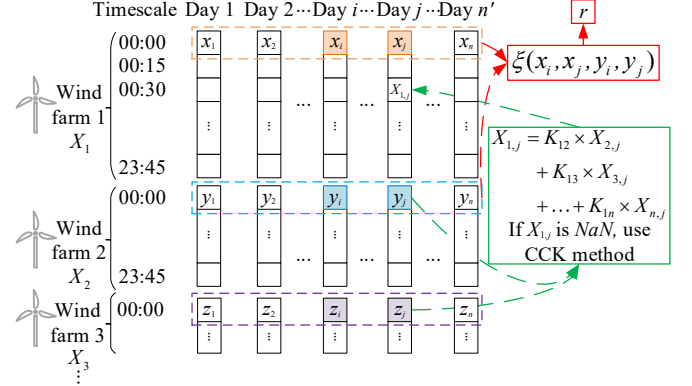

Fig. 3. Calculation of KRCC in different wind farms and the CPK method

In Fig. 3, a new compensating parameter that uses the KRCC (CPK), is defined as follows:

$$K_{\alpha\beta} = \frac{r_{\alpha\beta}}{\sum_{\lambda=1}^{n} r_{\alpha\lambda} - r_{\alpha\alpha}}, \alpha, \beta \in \{\alpha \neq \beta \mid \alpha, \beta = 1, 2, ..., n\} \qquad (15)$$

where $K$ stands for the CPK of any two wind farms (e.g. $\alpha$ stands for wind farm 1 and $\beta$ for farm 2) in the same cluster.

### C. Cluster Data Interpolation Method Based on KRCC (CCK)

One way to handle missing data in wind speed series is the moving average approach with autoregressive order (MAA), which is widely used in a single wind farm [31]. Firstly, the flags called "not a number (*NaN*)" are used to replace the missing values. Then, the *NaN* positions are patched by MAA as the average value of these preceding numbers of data with the assigned autoregressive order, which is defined as:

$$x_i = \frac{1}{l} \sum_{k=i-l}^{i-1} x_k \qquad (16)$$

where $x_i$ is the first missing data and $l$ is the autoregressive order. However, as the wind speed is not linear with time, MAA fails to fit it well.

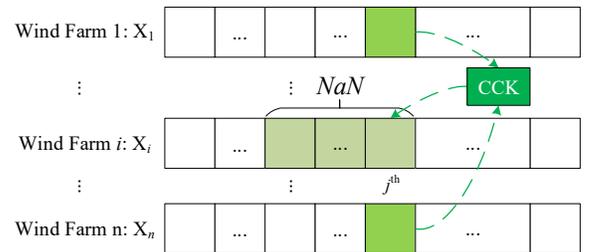

Fig. 4 Diagram of the CCK method

Furthermore, a multidimensional average method (MDAM)

is suggested to use wind cluster data to supplement the missing data [32]. However, further exploiting of correlation between the wind farm cluster is still a big challenge. In comparison, the wind farm Cluster data Corrective method based on CPK, abbreviated as CCK, is first proposed in this paper to estimate the missing wind speed value in wind farm cluster with CPK. The procedure of the CCK is demonstrated in Fig. 4, and the detailed calculation is available as follows:

$$X_{i,j} = K_{i1} \times X_{1,j} + K_{i2} \times X_{2,j} + \ldots + K_{in} \times X_{n,j} \quad (17)$$

where $X_{i,j}$ is the assumed *NaN* value in wind farm $i$, $X_{1,j}$, $X_{2,j}$, …, and $X_{n,j}$ is the wind speed data at the same time instants of the adjacent wind farms, while $K_{i1}$, $K_{i2}$ and $K_{in}$ denote the CPK between the target wind farm $i$ and the corresponding adjacent wind farms.

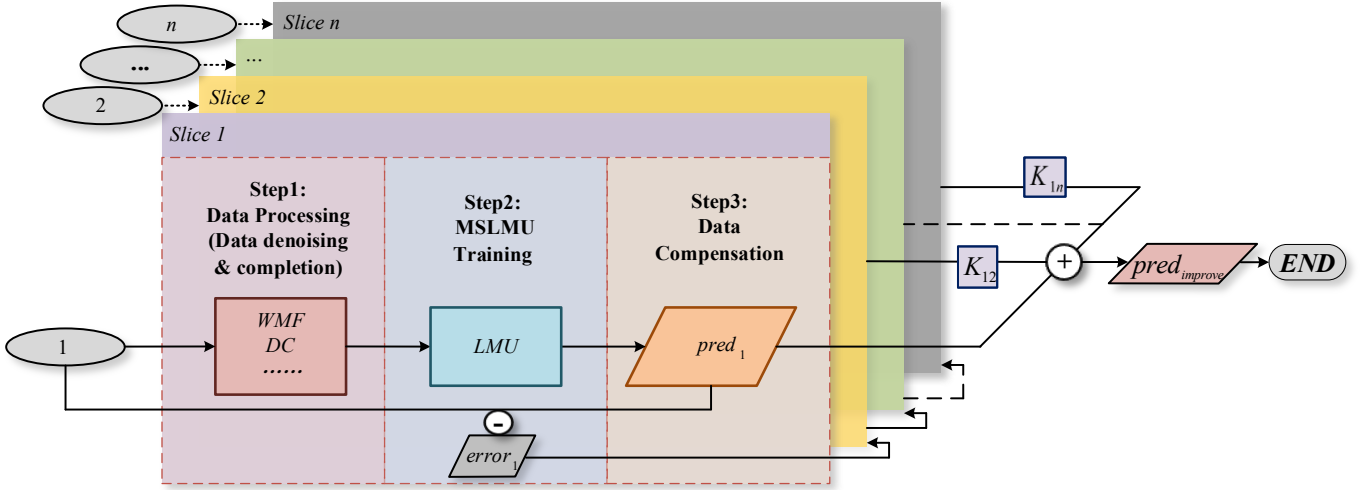

Fig. 5. The flowchart of the WMF-CPK-MSLMU model (take wind farm 1 as an example for a more convenient demonstration)

### D. The Strategy for WMF-CPK-MSLMU Modeling

Fig. 5 illustrates a newly proposed model for WMF-CPK-MSLMU that consists of $n$ slices and three steps.

WMF-CPK-MSLMU takes historical wind speed series in wind farm cluster as input and outputs an improved prediction for the target wind farm. The detailed steps of the proposed model WMF-CPK-MSLMU are given in what follows (take wind farm 1 as an example for a more convenient demonstration in Fig. 5 while without loss of generality):

***Step 1*** During the data cleaning preprocess, the original wind speed time series are collected, completed, denoised, and normalized, thus achieving the clean wind speed time series. Particularly, using the WMF method, redundant information is filtered from original data, thus leading to a wind speed series that is denoised. Moreover, whenever the speed of any wind farm is unavailable, the compensation model will make up for the continuous missing data by taking advantage of the wind speed of the adjacent wind farms.

***Step 2*** As described in Section II.B, the denoised time series are used to train the prediction model of the LMU neural network. Then it is applied for future prediction to produce the final results in WSF.

***Step 3*** Once the first slice of training is tested and finished, the test error is calculated as the prediction value of training for the next slice. The second and third slices can be trained and tested on this method, and the prediction results of each slice are obtained at the end.

Compared with WMF-MSLMU, WMF-CPK-MSLMU uses CPK to consider the spatial characteristics among wind speed series, which greatly improves the accuracy. Integrated with the above information, the prediction flow equation of the model can be constructed finally as follows:

$$pred_{improve} = pred_1 + \sum_{i=2}^{n} K_{1i} \times pred_i \quad (18)$$

where $pred_{improve}$ denotes the improved prediction value of the WSF in WMF-CPK-MSLMU for wind farm 1.

### E. Training Parameters

TABLE I
Characteristics of the ensemble model WMF-CPK-MSLMU

| Component models | Parameters | |
|---|---|---|
| WMF | Size | 5 |
| | Weights | [0.90, 0.381, 0.73, 0.66, 0.559] |
| CPK | Correlation weight of adjacent wind farms | |
| | site #1 to site #2 | 0.73~0.85 |
| | site #1 to site #3 | 0.58~0.67 |
| MSLMU | Slices | Configuration |
| | Slice 1 | Input shape: 10×1 |
| | | Unit number: 512 |
| | Slice 2 | Input shape: 5×1 |
| | | Unit number: 256 |
| | Slice 3 | Input shape: 5×1 |
| | | Unit number: 128 |
| | Activation: Sigmoid | |

TABLE I provides the parameters of the proposed ensemble model herein. In this paper, only three slices of LMU are chosen (let $n = 3$), while four or more slices will reduce the prediction accuracy and prolong the training time (details can be seen at Section IV.E). More generally, the proper value of $n$ for other
5

multi-wind farms can be selected according to the actual data and effect.

## IV. CASE STUDY

### A. Data Set

The subsequent evaluation of the proposed methodology is based on different data sets. These data are publicly available from the Bureau of Reclamation [33]. Considering that the correlation amongst multiple wind farms is proportional to the distance, three adjacent wind farms (AWF), site #1, site #2 and site #3, in the state of Oregon, USA (seen from the satellite map), are selected as the main test sites within the available range for this experiment. Site #1 and two non-adjacent wind farm (NWF) sites, site #4 and site #5, are employed as a comparison. More details can be found in TABLE II. In the following, the names of each wind farm are available in TABLE II, and the site index is used to represent station ID in this paper. The resolution of the raw data is every 15 minutes, and these observed samples (10000 samples are chosen) from each site are divided into three portions, of which, 70% is used as the training set, 20% is treated as the validation set, and the remaining 10% works as the testing set. For clarity, three data configurations are defined as: the normal set (complete data), the cluster set (multi-site data), and the single set (individual site only).

It should be noted in advance that the data presented in this section of the comparison experiments are obtained through the simulations on a personal computer equipped with Intel Core i7-8565U CPU and 16GB RAM, while the average values of ten replicate trials are used for analysis and discussion.

TABLE II
Geographic and statistical information of the datasets (from 1st January 2020 to 15th April 2020)

| Site Index | Station ID | State | Longitude and Latitude | Wind Speed (m/s) Mean | Standard Deviation |
|---|---|---|---|---|---|
| site #1 | MDXO | OR | (-121,45) | 4.76 | 4.49 |
| site #2 | MDNO | OR | (-121,45) | 4.43 | 4.02 |
| site #3 | MDSO | OR | (-121,45) | 3.75 | 3.89 |
| site #4 | CRSM | MT | (-114,48) | 3.49 | 2.65 |
| site #5 | BANO | OR | (-124,43) | 3.93 | 2.96 |

### B. Evaluation Indices

A total of six accuracy criteria are employed in this paper, including the mean absolute percentage error (MAPE), mean absolute error (MAE), and root mean square error (RMSE) (19 – 21) as well as their promotion percentages $P_M$ (22) [34].

$$MAPE = \frac{1}{T}\sum_{t=1}^{T}\left|\frac{wind_{real} - wind_{pred}}{wind_{real}}\right| \quad (19)$$

$$MAE = \frac{1}{T}\sum_{t=1}^{T}\left|wind_{real} - wind_{pred}\right| \quad (20)$$

$$RMSE = \sqrt{\frac{1}{T}\sum_{t=1}^{T}\left(wind_{real} - wind_{pred}\right)^2} \quad (21)$$

$$P_M = \frac{M_2 - M_1}{M_2} \times 100\%, M \text{ can be } MAPE, MAE, RMSE \quad (22)$$

where $T$ denotes the number of timestamps in the testing set, which is set to 1000 herein; $wind_{real}$ represents the real historical wind speed values and $wind_{pred}$ represents the predicted ones; $MAPE_1$, $MAE_1$, $RMSE_1$ are acquired from the proposed model WMF-CPK-MSLMU, while $MAPE_2$, $MAE_2$, $RMSE_2$ denote the evaluating indicators of other candidate models. $P_M<0$ means the evaluation index of model 1 is larger than model 2.

### C. Data Preprocessing

WMF is a powerful and promising noise reduction method, which can bring a good effect to depress the forecasting errors significantly, and when the predictor is fixed, WMF can always obtain the lowest MAEs combined with any one predictor in one-step forecasting [34]. Accordingly, the WMF is selected amongst various noise reduction methods in this paper, to depress the influence of white noise in the original data, thus improving the prediction accuracy [34].

Firstly, a set of comparative experiments on the forecasting accuracy between the non-denoised data and denoised data are built as shown in Fig. 6. When the LMU is used as a predictor, such denoised data can reduce MAPE, RMSE, and MAE by 87.25%, 64.92% and 62.99%, respectively, compared to the non-denoised data. Moreover, the effect of the denoising method combined with the MSLMU is similar (MAPE, RMSE, and MAE are reduced by 92.22%, 73.16%, and 73.55% in three-slice). Above all, the WMF can significantly contribute to the accuracy improvement in the experiment of WSF by adopting the LMU as the forecasting method (i.e., predictor).

Therefore, the denoising process with adopting the WMF is important for LMU to extract the essential features from wind speed time series and hence boost the forecasting accuracy.

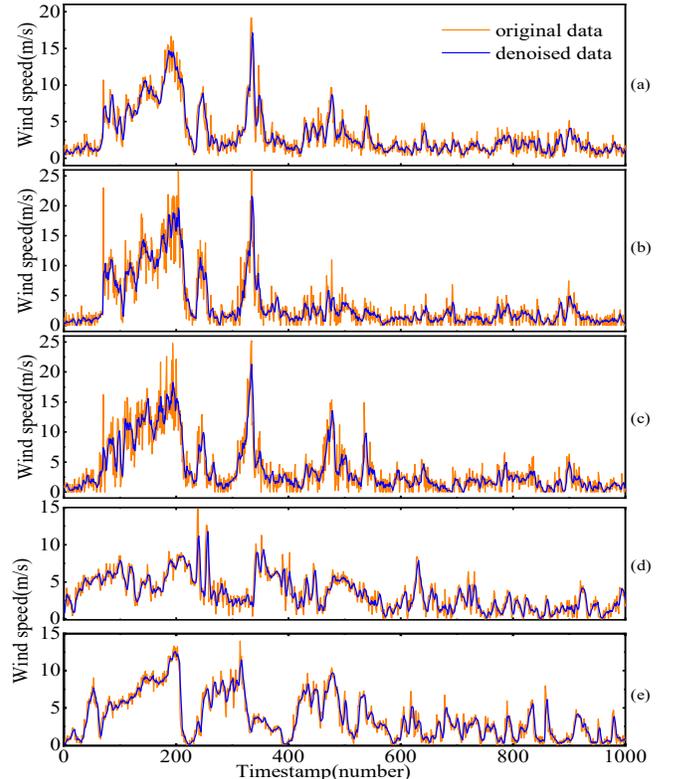

Fig. 6. Raw and denoised wind speed time series: (a) site #1 (MDXO); (b) site #2 (MDNO); (c) site #3 (MDSO); (d) site #4 (CRSM); (e) site #5 (BANO)

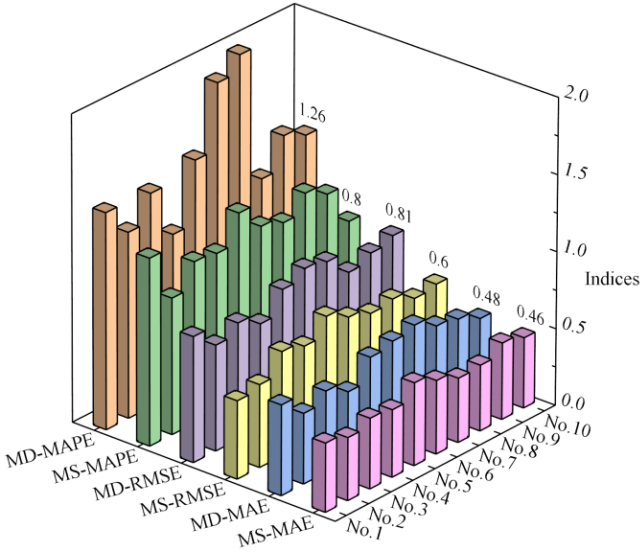

Fig. 7. Comparison of prediction accuracy between WMF-MDIS-LMU and WMF-CPK-MSLMU at site #1 through the repeated ten times experiment (MD represents WMF-MDIS-LMU; MS represents WMF-CPK-MSLMU)

TABLE III
Comparison of prediction CPU-time between WMF-MDIS-LMU and WMF-CPK-MSLMU at site #1

| Criteria | Training time (s) | | | Forecast (s) | Total (s) |
|---|---|---|---|---|---|
| | Max | Min | Avg | | |
| WMF-MDIS-LMU | **1830** | **1760** | **1790** | 1.23 | **1801** |
| WMF-CPK-MSLMU | 2760 | 2680 | 2720 | **1.01** | 2731 |

*D. Comparison on Different Strategies for the Wind Speed Prediction of Wind Farm Cluster*

The WMF-CPK-MSLMU strategy proposed in this paper is different from the one of reference [17], which presented a two-stage modeling strategy of extracting the spatial features firstly, followed by capturing the temporal dependencies amongst the extracted spatial features [17]. But on the contrary, the temporal relationship is established first, and then the spatial features and compensating errors can be extracted correspondingly for the WMF-CPK-MSLMU herein.

In Zhu et al. [17], multiple wind farms' historical wind and power data are jointly fed into the model as multichannel inputs, enabling the unified capture of spatial and temporal correlations. This structure inherently forms a Multi-Dimensional Input Set (MDIS). The comparison model is named as the weighted mean filter - multi-dimensional input set - Legendre memory unit (WMF-MDIS-LMU).

From the above Fig. 7, it can be concluded that the prediction accuracy of WMF-CPK-MSLMU model is better than WMF-MDIS-LMU. The mean values of MAPE, RMSE and MAE obtained by WMF-CPK-MSLMU in this experiment are 0.97%, 0.63 m/s, and 0.46 m/s, respectively. The three indices are 27.61%, 19.23%, and 16.36% lower than the ones obtained by WMF-MDIS-LMU. The test results highlight that the WMF-MDIS-LMU performs poorly in comparison to WMF-CPK-MSLMU due to its lack of a specific method for considering wind speed relationships amongst different wind farms.

In WSF, besides prediction accuracy, prediction efficiency is also a significant concern in practical applications. Based on the experiments performed on the same computer, TABLE III provides the total time consumption of these two strategies. According to TABLE III, WMF-CPK-MSLMU takes longer time for prediction due to the additional two-slice compensation structure, which is more complicated than WMF-MDIS-LMU. So WMF-CPK-MSLMU performs less efficiently compared to WMF-MDIS-LMU, but its accuracy has been greatly improved. In conclusion, both strategies for wind farm cluster prediction have pros and cons, and hence they can be chosen in different practical application scenarios.

*E. Comparison Among Different Numbers of Compensation Slices*

In-depth analysis is carried out to determine the optimal number of slices in the compensation model for prediction. In Fig. 8, for RMSE and MAE, the 3-slice model shows superior performance. RMSE and MAE values decrease, indicating a better accuracy. In contrast, models with 4 slices or 5 slices have relatively higher RMSE and MAE, suggesting reduced prediction accuracy and possible overfitting. The 2-slice model underperforms as well. The 3 slices model strikes a balance, offering improved prediction with manageable training time. This validates those 3 slices is the most suitable, while more slices may lead to reduced accuracy and longer training time.

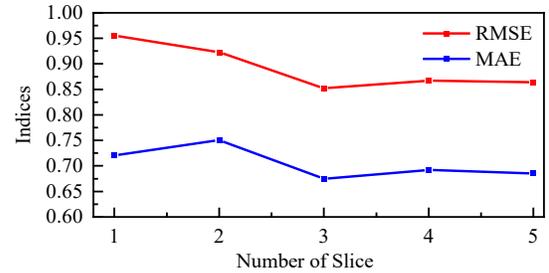

Fig. 8. Comparison among different number of compensation slices

TABLE IV
Comparison on the combined basis of five prediction algorithms in terms of single-slice and multi-slice prediction accuracy at site #1 *

| Prediction algorithm | | 1-slice | 2-slice | 3-slice |
|---|---|---|---|---|
| ELM | MAPE (%) | 1.08 | 1.13 | 1.05 |
| | RMSE(m/s) | 0.72 | 0.74 | 0.73 |
| | MAE (m/s) | 0.51 | 0.52 | 0.51 |
| SRNN | MAPE (%) | 1.44 | 1.21 | 1.20 |
| | RMSE(m/s) | 0.77 | 0.68 | 0.67 |
| | MAE (m/s) | 0.60 | 0.53 | 0.52 |
| LSTM | MAPE (%) | 1.12 | 0.99 | 0.98 |
| | RMSE(m/s) | 0.65 | 0.58 | 0.57 |
| | MAE (m/s) | 0.52 | 0.45 | 0.44 |
| GRU | MAPE (%) | **1.03** | 0.96 | 0.95 |
| | RMSE(m/s) | **0.61** | 0.57 | 0.56 |
| | MAE (m/s) | **0.47** | **0.44** | 0.43 |
| LMU | MAPE (%) | 1.20 | **0.94** | **0.93** |
| | RMSE(m/s) | 0.70 | **0.56** | **0.55** |
| | MAE (m/s) | 0.57 | **0.44** | **0.42** |

\* The minimum prediction errors are in bold.

*F. Comparison of Different Methods for the Wind Speed Prediction of Wind Farm Cluster*

To verify the forecasting performance and compensation effect of different prediction methods with WMF, a comparison experiment combining with different prediction algorithms and different compensation slices is established. The experimental results are shown in TABLE IV. Overall, 10 replicate trials are used to obtain the average values for analysis and discussion.

The prediction algorithms selected for comparison on the experiments of this subsection have been applied widely in the field of single WSF.

First, it can be seen from TABLE IV that amongst the five different prediction algorithms, GRU exhibits better prediction accuracy with the one-slice model which is superior in the uncompensated single wind farm prediction. In detail, the MAPE, RMSE, and MAE obtained by GRU are 1.03 %, 0.61 m/s, and 0.47 m/s, individually. Moreover, the prediction accuracy of most of the candidate algorithms improved with the increase in the number of compensation slices. Amongst them, LMU shows its strong adaptability to the compensation slice structure in particular. With the three-slice compensation structure, the error metrics MAPE, RMSE, and MAE of LMU drop to 0.93 %, 0.55 m/s, and 0.42 m/s, respectively.

Fig. 9 demonstrates the improvement of LMU compared to the four promising prediction methods in terms of different error metrics. Candidate algorithms, ELM, SRNN, LSTM and GRU are all applied with the same modeling strategy of LMU. Taking WMF-CPK-MSELM for example, it is named in a similar way with WMF-CPK-MSLMU. As shown in TABLE IV, LMU performs better in terms of RMSE under the three-slice structure and has relatively small numerical values for MAPE and MAE. Therefore, with the fixed noise reduction method WMF, the compensation structure of the multiple wind farms leads to a significant improvement in the accuracy of WSF. Meanwhile, considering the three indices together, the LMU possesses a strong adaptability and higher prediction accuracy under the multi-slice compensation based on the data of wind farm cluster. Meanwhile, Fig. 9 presents the fitting effect of these five prediction algorithms in terms of the prediction performance. It can be investigated that all the prediction algorithms selected for the experiment have the fitting ability. Amongst them, the deviations of ELM and SRNN prediction curves are larger compared with LSTM, GRU, and LMU. Moreover, the fitting effect of some local parts is improved as the number of slices increases.

Second, the improvement on prediction accuracy of some different prediction methods by the compensation structure is quantified. TABLE V shows the accuracy improvement of the multi-slice structure for the five methods compared to 1-slice - the single wind farm prediction. Through quantifying the percentages and comparing them with different slices, it can be found that the improvement effect of three slices compared to two slices is mostly achievable. Comparing with five different methods, it is clear that the compensation structure makes the most significant effect on the accuracy improvement of LMU, with 22.83%, 20.20%, and 23.42% of decrease for MAPE, RMSE, and MAE, respectively, under the three-slice structure.

TABLE V
Effectiveness of compensation slices on the promotion percentage of evaluation indices based on 1-slice respectively at site #1 *

| Prediction algorithm | | 2-slice | 3-slice |
|---|---|---|---|
| ELM | $P_{MAPE}$ | -4.96% | 2.78% |
| | $P_{RMSE}$ | -2.55% | -0.75% |
| | $P_{MAE}$ | -2.16% | -0.57% |
| SRNN | $P_{MAPE}$ | 15.88% | 16.15% |
| | $P_{RMSE}$ | 11.41% | 11.81% |
| | $P_{MAE}$ | 12.80% | 13.17% |
| LSTM | $P_{MAPE}$ | 12.06% | 12.72% |
| | $P_{RMSE}$ | 10.33% | 11.40% |
| | $P_{MAE}$ | 13.28% | 14.40% |
| GRU | $P_{MAPE}$ | 6.40% | 6.87% |
| | $P_{RMSE}$ | 5.73% | 6.90% |
| | $P_{MAE}$ | 7.83% | 8.72% |
| LMU | $P_{MAPE}$ | **22.09%** | **22.83%** |
| | $P_{RMSE}$ | **18.91%** | **20.20%** |
| | $P_{MAE}$ | **22.44%** | **23.42%** |

* The maximum improvement ratio is in bold, while a negative number means the error index becomes even larger.

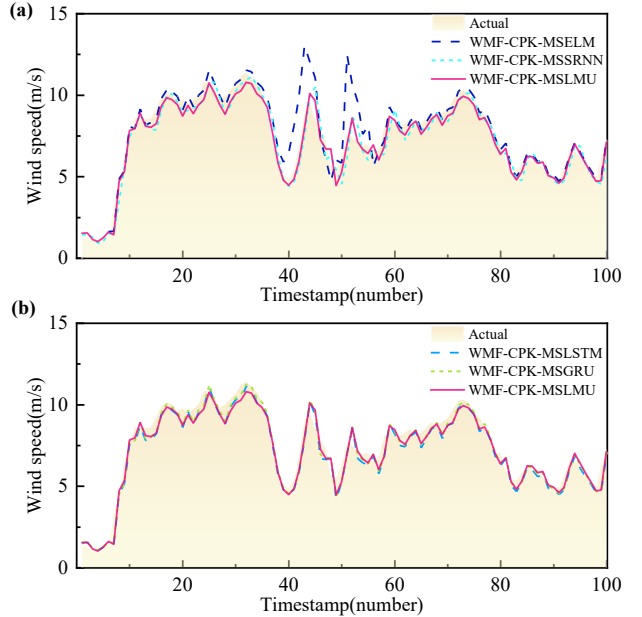

Fig. 9. Comparison of 3-slice WSF results at site #1 by five approaches: (a) results for WMF-CPK-MSELM/MSSRNN/MSLMU; (b) results for WMF-CPK-MSLSTM/MSGRU/MSLMU

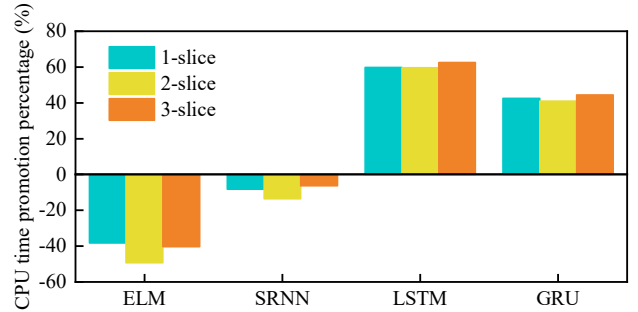

Fig. 10. CPU times promotion percentages of four candidate models by WMF-CPK-MSLMU

Apart from prediction accuracy, prediction efficiency is also another critical concern for the WSF in practical applications. Conducted based on the same personal computer, Fig. 10 shows the total prediction time required by some different prediction algorithms under different slices. Besides, SRNN and ELM have slightly faster prediction speeds than LMU, while LSTM and GRU possess much slower prediction speeds than LMU. Among these deep learning methods, according to Fig. 10, the average prediction time of LMU is 62.58% less than that of LSTM, and 44.4% less than that of GRU. Overall, the prediction efficiency of LMU is in the middle level among the five prediction methods.

In conclusion, compared to the other four algorithms which are commonly used for WSF, the WMF-CPK-MSLMU model proposed in this study has a higher prediction accuracy and also a high prediction efficiency in WSF based on the wind farm cluster data. The LMU outperforms RNNs that utilize memory cells and gating mechanisms like GRU and LSTM in memory capacity and handling long-term temporal dependencies while mitigating issues such as unstable gradients and saturation effects. Adoption of LMU in a multi-slice compensation model overcomes its limitations in single-slice accuracy while fully leveraging its efficiency and large data processing capacity to produce the improved results for WMF-CPK-MSLMU.

TABLE VI
Comparison of different methods for data supplement

|  | Normal | Single | Comparison between normal and single ($P_M$) | Cluster | Comparison between normal and cluster ($P_M$) | Improvement from single to cluster ($P_M$) |
|---|---|---|---|---|---|---|
| 5 missing values |  |  |  |  |  |  |
| MAPE | 0.086 | 1.599 | -1769.62% | 0.647 | -86.78% | 59.54% |
| RMSE | 0.905 | 1.594 | -76.03% | 1.080 | -16.20% | 32.21% |
| MAE | 0.527 | 1.281 | -142.89% | 0.931 | -43.34% | 27.34% |
| 10 missing values |  |  |  |  |  |  |
| MAPE | 0.048 | 4.064 | -8283.61% | 0.353 | -86.25% | 91.33% |
| RMSE | 0.822 | 3.237 | -293.68% | 1.062 | -22.55% | 67.20% |
| MAE | 0.567 | 2.670 | -370.81% | 0.867 | -34.59% | 67.53% |
| 15 missing values |  |  |  |  |  |  |
| MAPE | 0.035 | 4.109 | -11806.43% | 0.228 | -84.88% | 94.44% |
| RMSE | 0.796 | 3.476 | -336.82% | 1.052 | -24.37% | 69.73% |
| MAE | 0.607 | 2.986 | -391.51% | 0.904 | -32.82% | 69.71% |

*G. Short-Term WSF with Missing Data*

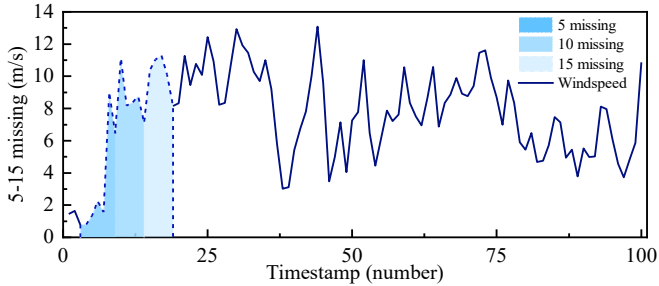

Fig. 11. Demonstration of wind speed time series with several data missing at site #1

The aforementioned experiments are based on normal wind speed data, but a few anomalies are often encountered. Since occasionally there are many missing values exist in the original data files, appropriate methods must be used to process these data and hence ensure the robustness of the predictions. Two compensation methods, MAA in a single set and CCK in a cluster set, are selected for comparison as follows.

In the experiments, the wind speed data in the testing set at site #1 is presented, and the wind speed data with 5, 10, and 15 continuous missing values, respectively, are chosen to testify the predictive accuracy. It is noted that the values of all evaluation indicators are calculated within the range of missing data. Fig. 11 depicts the detailed time interval according to the tendency of wind speed data, while the results can be seen in TABLE VI, which indicates that the CCK method is relatively more reliable than the MAA method. TABLE VI compares three sets of wind speed data, among which are complete set without missing data (normal), set with MAA method for missing data (single), and set with the CCK method for missing data (cluster). In TABLE VI, as the number of missing data increases, $P_{MAPE}$, $P_{RMSE}$ and $P_{MAE}$ between the cluster set and normal set reach -86.78%~-84.88%, -24.37%~-16.20%, and -43.34%~-32.82%, separately, while the three indices between single set and normal set can reach to -11806.43%~-1769.62%, -336.82%~-76.03%, and -391.51%~-142.89%, separately.

Evaluation indices in both cluster set and single set get worse when the missing data becomes more serious, but the cluster set significantly inhibits the fluctuation. Moreover, the promotion percentage $P_{MAPE}$, $P_{RMSE}$ and $P_{MAE}$ between the cluster set and single set reaches to 59.54%~94.44%, 32.21%~69.73%, and 27.34%~69.71%, individually.

It is easily observed in TABLE VI that the method which considers the correlation of wind farm clusters can remarkably improve the forecasting accuracy, compared to the method in the single wind farm. In conclusion, the CCK method possesses a superior effect in improving the robustness and accuracy of WSF based on the data of wind farm clusters.

V. CONCLUSION

In this paper, a new ensemble model WMF-CPK-MSLMU is presented. Firstly, as an effective noise reduction method that significantly decreases the forecasting errors and improves the prediction accuracy by depressing the white noise influence, WMF is used in WSF herein. Then, with the introduction of CPK, the robustness of proposed model has been significantly enhanced. Furthermore, the MSLMU is firstly used in the WSF of wind farm clusters, which achieves a good balance between accuracy and efficiency.

The LMU employs a continuous-time linear dynamical system that projects input history onto a set of shifted Legendre polynomials via coupled ODEs. This yields a fixed-size, orthogonal, and scale-invariant representation of the past— capable of preserving temporal dependencies spanning tens of thousands of time steps. By contrast, GRU and LSTM rely on nonlinear gating that can saturate and lose temporal resolution over long sequences. The LMU's mathematically grounded delay network preserves fine-grained temporal dynamics, adapts its effective time step during training, and avoids gradient instability. These features align closely with wind-farm cluster data, which exhibit multiscale temporal variability (e.g., diurnal cycles, turbulence bursts) and require stable, long-horizon forecasting.

Based on test results, conclusions can be drawn as follows.

1) Compared to the traditional wind farm cluster prediction method of constructing the input sets, WMF-CPK-MSLMU can achieve higher accuracy and better robustness. The promotion percentage in MAPE, RMSE, and MAE by means of WMF-CPK-MSLMU reaches 16.36%~27.61%.

2) Amongst the promising deep learning models for WSF, WMF-CPK-MSLMU is both fast and accurate. In particular, it reduces MAPE by 14.70%~15.96% compared to WMF-CPK-MSGRU, and by 8.80%~10.11% in comparison to WMF-CPK-MSLSTM. Furthermore, WMF-CPK-MSLMU works much faster than the two algorithms by 62.58% and 44.40%, separately.

3) When dealing with the loss of continuous data, WMF-CPK-MSLMU supplements it well with the wind speed data of AWFs, and the whole model has a good robustness. The $P_{RMSE}$ between MAA and CCK reaches 32.21%~69.73%.